\documentclass[twoside,11pt]{article}

\usepackage{dnd}
\usepackage{times}
\usepackage{multicol}

\begin{document}

\title{Report from the NSF Future Directions Workshop, Toward User-Oriented Agents: Research Directions and Challenges}

\author{\name Maxine Eskenazi \email max@cs.cmu.edu \\
       \addr Carnegie Mellon University 
       \AND
       \name Tiancheng Zhao \email tianchez@soco.ai \\
       \addr SOCO.AI}

\maketitle
\begin{abstract}%
This USER Workshop was convened with the goal of defining future research directions for the burgeoning intelligent agent research community and to communicate them to the National Science Foundation. It took place in Pittsburgh Pennsylvania on October 24 and 25, 2019 and was sponsored by National Science Foundation Grant Number IIS-1934222. Any opinions, findings and conclusions or future directions expressed in this document are those of the authors and do not necessarily reflect the views of the National Science Foundation. The 27 participants presented their individual research interests and their personal research goals. In the breakout sessions that followed, the participants defined the main research areas within the domain of intelligent agents and they discussed the major future directions that the research in each area of this domain should take. 
\end{abstract}

\begin{center} \bf 2019 Toward User-Oriented Agents Workshop participants/report contributors: \par \end{center}

\begin{multicols}{2}
    \begin{itemize}
        \item Prabhakaran Balakrishnan, National Science Foundation
        \item Jeffrey Bigham, Carnegie Mellon University
        \item Carlos Busso, University of Texas Dallas
        \item Jamie Callan, Carnegie Mellon University
        \item Vivian Chen, National Taiwan University
        \item Barbara DiEugenio, University of Illinois at Chicago
        \item Maxine Eskenazi, Carnegie Mellon University
        \item Milica Gasic, Heinrich Heine University
        \item Joakim Gustafsson, KTH Royal Institute of Technology
        \item Julia Hirschberg, Columbia University
        \item Rebecca Hwa, National Science Foundation
        \item Andruid Kerne, National Science Foundation
        \item Tanya Korelsky, National Science Foundation
        \item Diane Litman, University of Pittsburgh
        \item Shikib Mehri, Carnegie Mellon University
        \item Roger Moore, University of Sheffield
        \item LP Morency, Carnegie Mellon University
        \item Mari Ostendorf, University of Washington
        \item Antoine Raux, Apple
        \item Verena Reiser, Heriot Watt University
        \item Wade Shen, Actuate Innovation
        \item Amanda Stent, Bloomberg
        \item David Suendermann-Oeft, Modality.AI
        \item David Traum, University of Southern California
        \item Stefan Ultes, Daimler
        \item Clare Voss, Army Research Laboratories
        \item Marilyn Walker, University of Santa Cruz
        \item Tiancheng Zhao, SOCO.AI
    \end{itemize}
    \end{multicols}

\section{Introduction}

Intelligent agents have entered the realm of popular use. Over 50,000,000 agents are present in homes in the US alone. As they become part of the household, the developing dependencies and immense opportunities challenge the AI community to expand its research horizons. While much past research has centered on the intelligent agent itself, future research must center on the user. When research only develops the qualities of the agent and ignores the needs and individual characteristics of users, agents perform poorly. Poor performance leads to the user’s loss of faith. Not only will the user no longer use an agent, but loss of confidence and public interest in using agents can have serious consequences for research. This can be as harmful to research as the drastically reduced funding that automatic speech recognition (ASR) experienced in the 1990s. The agent must serve the needs of the user~\cite{eskenazi2019beyond}. Intelligent agent research should center around the vision of an interlocutor that converses easily with one or more users on a large variety of topics in many diverse settings. It is a flexible, robust partner in many everyday endeavors.

In the early days of dialog research, the user was an equal partner with the agent. Assessment was often based on the PARADISE approach~\cite{walker1997paradise}. Researchers measured the length of dialogs and their success rate. At the time, this approach aimed at measuring whether the user obtained what they had requested. Speed and correct information were considered to be the characteristics that would satisfy the user. This was confirmed with user feedback surveys. Since that time, researchers have examined new aspects of the agent, for example, \textit{emotion}~\cite{gratch2001tears}, concentrating on whether the system chose to express the “right” emotion given some context where “right” is a match with something that occurred in the past context. Then \textit{rapport} and other affective characteristics were added to the agent~\cite{gratch2007creating}. Again, the focus was on whether the \textit{agent} was expressing this correctly, given past context. The academic community has experienced difficulty in obtaining data from real users. Consequently, researchers have turned to simulated users and paid users. Simulated users can generate a significant amount of data, but the utterances generated can never deal with elements that were not in the dataset they were trained on. It is also known that paid users respond very differently from real ones. \cite{ai2007comparing}, for example, showed that paid users accept undesirable information from a system simply because getting what they wanted would take more time. Personae have been created. The ConvAI2 assessment ~\cite{dinan2019second} determined whether the correct persona was chosen based on past context but not on the agent’s usefulness. Present measures used to determine an agent's success often have nothing to do with the user. For example, the BLEU~\citep{papineni2002bleu} lexical similarity score has little to do with dialog success~\cite{liu2016not}. In sum, much state-of-the-art intelligent agent research only centers on and assesses what an agent produces, not what the user has obtained from their conversation. A few exceptions do exist, for example, the agents, such as the twin museum guides and the agent to diagnose mental illness produced by researchers at the University of Southern California (USC)~\cite{traum2008talking}.

%%%%
\subsection{The role of the agent}
One of the overarching goals of the intelligent agent community has been to test whether the agent can be designed to be indistinguishable from a human. This is often referred to as the Turing test~\cite{turing1950computing}. The Turing test presented the very seductive idea of having a machine deceive a human. Much research every year goes into creating an agent capable of making the user believe that it is human. And some research has gone as far as to produce agents that reside in the Uncanny Valley~\cite{moore2012bayesian}. Deception has created fear in the general public. Consequently much “explaining” has taken place (“robots are not out to take over the world, they will not replace humans”~\cite{kflee2019}). 

This USER Workshop report relays the intelligent agent research community’s vision of the future directions it should follow which government funding agencies should encourage. The vision builds on the relationship between the agent and the user rather than on the appearance and functions of the agent alone. It has a well-defined set of goals that lead to clear directions for research and for assessing progress. User satisfaction and self-assessment of agent usefulness are pillars in defining and assessing the role of the user as well as in determining how well agents serve their users. At the same time, areas related to user wellbeing such as health, behavior and education have their own objectives, expert measurements such as blood pressure or post-test improvement.

The USER Workshop participants agreed that the intelligent agent’s main role is to be the user’s partner. The agent should work hand in hand with the user toward some mutual goal. This implies that every type of agent should have a goal, for example, to serve information or just provide companionship. With this role in mind, we measure success from our own expert observation, from third party observation and, especially, from the user’s point of view. Examples of seamless partners are: a pen is the partner of the hand when writing; a car is the partner of a rider when traveling. The iPhone has taken the place of a watch as a partner in telling time. Apple store employees ask customers to come back after an hour to pick up a phone left for a simple repair. They observe that most customers come back in 15 minutes and are surprised to find that it has not been an hour. They appear to have lost the ability to estimate time without their iPhone. 

If the agent traditionally begins a dialog with, “How may I help you?”, it should end it with “Have I helped you?”.

\subsection{Present research in user-oriented intelligent agents}

Some research has taken the user into account. PARADISE~\cite{walker1997paradise}, which used task completion while optimizing for the number of utterances, is a measure that is still employed in many publications.

\newcite{tan2018multimodal} used LSTMs to predict the empathetic response of the user over time, concentrating on the emotions of the user rather than those of the agent. \newcite{muralidharanleveraging} sped up data annotation of user engagement by using specific signals from the user that reflect their engagement. They then used multitask learning and validation to increase labelling accuracy. They chose to create coarse grain labels, which are easier to obtain and more reliable and then used them to generate finer grain labels. \newcite{hancock2019} used a configuration that enabled the user to give the agent immediate feedback when it made an error. The agent first detected that the user was not satisfied and then generated an appropriate question so that the user could explain what should have been said. They then modeled the user’s answer to generate future utterances. \newcite{eskenazi2019beyond} used previous context and the user response immediately following the agent utterance being assessed to determine the quality of that utterance. This shows better results than past work that only used the past context, ignoring the user’s reaction to the agent utterance.

One way to carry out intelligent agent assessment is to employ third party human judgment. \newcite{lowe2017towards} determined the validity of the agent’s response based on criteria such as topicality, informativeness and whether the context required that the background information be understandable. \newcite{shah2016interactive} demonstrated the possibility of interactively adapting to the user’s response using reinforcement learning so that the system developer could indicate the desired end states. This made the agent’s speech more varied than the speech in the training database. By actively modifying the agent using the simulated user turn that follows a given system turn, the new system response became increasingly more appropriate and useful to the user. 
 
\subsection{Impacts of user-oriented intelligent agent research}
\subsubsection{Applications that influence many}
The impact of intelligent agent research has gone beyond the scientific community, where novel techniques and massive use of neural networks abound. Intelligent agents are now a part of everyday life. They serve general needs for information and chat, as we see in the most popular applications: SIRI (Apple.com), Amazon Echo (Amazon.com), etc. Agents are also used in more targeted domains. Some examples are list below. See Section~\ref{sec:app} for more details.
\begin{enumerate}
    \item Seniors: An intelligent agent for seniors that is more understandable and gives out information in a more memorable manner~\cite{mehri2019cmu}
    \item Healthcare: An agent that interviews patients to determine mental health issues~\cite{rizzo2016autonomous}. An agent that helps with home care for patients with dementia~\cite{cailleteau2019} 
    \item Education: agents that are used to assess non-natives’ levels of fluency~\cite{litman2016towards}. Tutoring in the domain of Machine Learning using dialog ~\cite{kojima2019}. Education on sexual assault and harassment~\cite{artstein_digital_2019} 
\end{enumerate}

\subsection{Transformative infrastructure for industry in the next 10 years}

In industry research centers on agents that serve the company’s purpose. For example, Alexa ~\cite{ram2018conversational} often helps Amazon sell merchandise. Advances in industrial agents benefit from a significant amount of data from real users and superior computing power. Since the advances in industry are usually driven by commercial goals, they do not address much of the fundamental research needed to make transformational advances. Neural networks are the state of the art in intelligent agent architecture research. Despite claims that they are used in commercial systems, most companies continue to employ droves of workers who create handwritten rules rather than rely on the nascent and very rigid neural networks. In the coming years, industry is poised to implement the more reliable neural systems that will come from academic research.
 
\subsection{A healthy research community}

A healthy research community is a community that has diverse research directions and methodologies. Encouraging research on user-centric intelligent agents can create a balance with present agent-centric research. Multidisciplinarity is another feature of a thriving community. It provides opportunities for future researchers to continuously innovate by adapting ideas from different disciplines. Additionally, user-centric intelligent agents may provide answers to one of the hardest challenges for current state-of-the-art systems: system evaluation and learning with the human-in-the-loop.

\section{Research Topics \& Future Directions}
The USER Workshop participants defined seven broad areas of future intelligent agent research: applications; dynamic views of user-agent interactions; infrastructure; intelligent agents as good actors; multimodal, grounded and situated interaction; robust and flexible dialog management; data reduction. Six crosscutting themes have surfaced that reflect issues common to several areas. These themes are: user-centric focus, multidisciplinary research teams, dialog conditions and situations, focused shared tasks, ethics and privacy, architecture. Table 1 shows the intersections of the themes and the broad areas.

\begin{table}[h]
\small
\begin{tabular}{|p{0.125\textwidth}|p{0.12\textwidth}|p{0.12\textwidth}|p{0.12\textwidth}|p{0.12\textwidth}|p{0.12\textwidth}|p{0.12\textwidth}|}
\hline
Broad area/Theme                             & User centric & Architecture & Multi disciplinary & Conditions \& situations & Shared tasks & Ethics \& privacy \\ \hline
Applications                                 & x            &              & x                 & x                        &              &                   \\ \hline
Dynamic user-agent interactions              & x            & x            & x                 &                          & x            &                   \\ \hline
Infrastructure                               & x            & x            &                   &                         & x           &                   \\ \hline
Intelligent agents / good actors             & x            &              &                   &                          &              & x               \\ \hline
Multimodal, grounded \& situated interaction & x            & x            & x                 & x                        & x            &  x               \\ \hline
Robust \& flexible dialog management         & x            & x            &  x               &  x                     &              &                   \\ \hline
Data reduction                               & x            & x          &                   &                          &   x         &                   \\ \hline
\end{tabular}
\caption{Broad areas (rows) and themes (columns)}
\end{table}

%\subsection{Broad areas}
%The workshop participants agreed on the seven broad areas mentioned above. The reasoning for each area follows.
%TEXT TO COME FROM BREAKOUT TEXTS

\subsection{Crosscutting Themes}
As described in detail below, each area discussion revealed a set of future directions that USER participants believed should be encouraged in the coming years. 

 Before describing the areas of research, this section discusses the crosscutting themes.  

\subsubsection{User-centric inspiration}
Researchers need to understand the user’s goals and beliefs in order to orient their research toward the user. One way to achieve this is to study human dialog. Another is to assess using measures such as user satisfaction that clearly reflect the success from the user's point of view.

\subsubsection{Multidisciplinary research teams}
In order to create truly useful intelligent agent applications and to understand intrinsic aspects of dialog, research teams should include experts from multiple domains. Some domains that are most relevant are linguistics, pragmatics, reasoning and psychology. Groups should also include area experts relevant to the type of agent they are creating. for example, individuals specialized in healthcare or education if the application is on that topic.

\subsubsection{Conditions and situations}
Intelligent agents have the potential to be present in a large variety of situations. They will take part in multiparty dialogs where the parties are a combination of humans and agents. They will be endowed with a broad range of modalities, including prosody and haptics. Situated dialog will find them pointing to and handing over physical objects and detecting/acting on objects in real situations. They will modify their style of delivery to support a variety of user preferences, such as verbosity. Thus they will become more personalized and dynamic. They will interact with users for longer conversations that will take place more and more frequently.

\subsubsection{Focused shared tasks}
The intelligent agent community increasingly assesses its progress through shared tasks. Some of the types of tasks that the USER Workshop participants envisaged are: situated dialogs, multiple thread dialogs involving complex planning, learning with less data and multimodal dialog.

\subsubsection{Ethics and privacy}
Although this theme is a broad area in itself, several other areas also mentioned it. Going forward, the design of research projects should include dealing with misuse and adversarial attacks. There should be concerted efforts in the area of privacy-preserving speech/visual/language processing and machine learning. The tension that exists between preserving privacy and having access to user models and audio-visual signals that can improve agent performance must be addressed. And finally, funders should favor proposals that include the creation of education and outreach programs on ethical issues.

\subsubsection{Architecture}
Although intelligent agent architecture has rapidly changed over the past few years, if agents are to realize their full potential, it is important that newcomers be able to easily join the research community. For this, plug-and-play systems that afford easier entry to the field must be developed. These systems can be treated as end-to-end black boxes. This will enable new researchers to come up to speed without prior knowledge of the workings of each module. 

Incremental speech processing must be generalized toward incremental multimodality. Researchers must combine structured representations that support inference with neural models (unstructured text). Explicit representations of dialog that can be shared across domains, tasks and applications should be favored. The community also needs a repository of pre-trained models accompanied by a usage guideline.

The following sections provide a detailed description of the seven areas of future research that were strongly advocated by the USER Workshop attendees.  
 
% SECTION 1: APPLICATION (DAVID SUENDERMANN)
\section{Applications}
\label{sec:app}
\subsection{Introduction}
Taking the user into account is essential in intelligent agent research across a multitude of applications. While all agents have their own merits, publicly funded research on user-oriented agents should follow a number of guiding principles:
\begin{enumerate}
    \item Applications should have three purposes: (a) advance fundamental research, (b) provide opportunities for commercialization, (c) do good for society
    \item Applications should be different from those that have already been broadly addressed in the past decades of intelligent agent research (flight booking, restaurant reservation, bus scheduling)
    \item Application teams should foster interdisciplinary relationships with scientific fields outside of computer science
\end{enumerate}

Although there are many relevant fields that satisfy these principles, two broad fields are:
\begin{itemize}
    \item Healthcare applications, e.g. diagnostics agents, caretaking agents, wellness agents, scribe agents, rehabilitation agents, and so on
    \item Educational applications, e.g. language learning agents, teacher training agents, interviewer agents, sourcing agents, assessment agents, tutoring agents
\end{itemize}

This section contains a detailed discussion of these two application areas and poses several key research questions related to them It also discusses the importance of interdisciplinarity in user-oriented agent research, and formulates major directions for intelligent agent applications.

\subsection{Healthcare}
Intelligent agents give healthcare providers new means of conducting intake interviews and training. This technology is particularly relevant for people who find it difficult to consult a healthcare provider. For example, military veterans returning from deployment may feel a stigma around mental health, and especially around post-traumatic stress disorder (PTSD)~\cite{rizzo2016autonomous}.  Intelligent agents acting as virtual interviewers can offer an alternative to a first meeting with a human for people feeling uncertain about mental health. In a study by~\newcite{lucas2014s}, people who were told they were interacting with a fully-automatic computer agent (as opposed to a human operating the agent) felt lower fear of self-disclosure, lower impression management, displayed their sadness more intensely, and were rated by observers as more willing to disclose. These virtual interviewers can be extended to become decision support tools for clinicians and healthcare providers by adding the capability of automatically detecting behavior markers of mental health~\cite{devault_simsensei_2014}. Behavior markers can be detected from multiple modalities: language (what is said), acoustics (how it is said) and visuals (expressions and gestures). Examples of behavior markers discovered using multimodal AI technologies include smile dynamics as a marker of depression~\cite{scherer2014automatic}, speech disfluencies as marker of psychosis~\cite{vail2018toward} and voice quality as marker of suicidal ideation~\cite{venek2017adolescent}. Multimodal behavior markers offer clinicians objective measures of the patient’s behavior that are relevant to their assessment of mental health symptoms and disorders. Intelligent agents can also be used as training tools for junior clinicians or for healthcare workers. An example is the Virtual Patient system which can create a safe environment for clinicians-in-training to interact with different types of patients to improve their assessment skills~\cite{kenny2007virtual}.

\subsection{Education}
Intelligent agents have been used to support a variety of interactive educational applications in a cost-effective and scalable manner. In one-on-one settings where the goal is to increase students’ knowledge about a particular domain (often STEM) via conversational interaction, agents have been developed that can take the role of tutor, peer, or tutee~\cite{graesser2005autotutor,kersey2009ksc,lubold2018automated}, reflecting different theories from the learning sciences about how to foster student learning via dialog. It has also been hypothesized that agents that are more user-oriented and that can detect and adapt to student affective states (e.g.,~\cite{litman2014evaluating}) or that can increase rapport~\cite{zhao2018socially} will be even more effective at increasing student learning. 

In the area of language learning and assessment, having students converse with intelligent agents has been proposed as a method for supporting both formative and summrative tasks (e.g., ~\cite{litman2016towards,evanini2018game,ramanarayanan2018toward}).  Finally, there have been efforts to move beyond one-on-one computer-student conversational interactions, for example, by enabling trialogues between a student and two agents, by facilitating a student’s dialog with other human students as in computer-supported collaborative learning, and by enabling students to observe the training dialogs of other students and/or intelligent agents (e.g.; 2019~\cite{kojima2019,kumar2014triggering,piwek2007t2d}). With respect to supporting teachers and others in the workplace rather than students, intelligent agents can support job training by simulating participation in target scenarios (e.g., sexual assault education by interacting with victim agents~\cite{artstein_digital_2019}.  Thanks to the challenges that “off the shelf” agents often face when applied to educational problems and data, due to differences in evaluation criteria, user modeling needs, dialogue structures, etc., this socially important application area typically produces innovative research.

\subsection{Other Application Areas} 
Other applications that spring from the agent-user partnership are, for example, factory assistants and aides for seniors.

Factory workers use agents for a variety of narrowly–defined applications. Workers inspecting equipment use agents to record their observations. These agents must adapt to the user’s background and preferences. For example, some employees have had the same job for decades and need very little help. The agent in this case just provides hands-free data entry, and the worker may only need one dialog turn to get something done. A newer employee, on the other hand, needs more help, implicit training from a chattier agent, until they are familiar with the task. Some workers who have been on the job for a while may still need some help. An agent should be able to detect that. It should also know when the user needs some extra time between turns. Here, assessment determines whether the worker was able to correctly finish their task, if they did it with less error, if they felt free to ask for extra information when needed and if they would use the agent again.

For seniors getting transportation information, agents should take into account the fact that human capabilities change as we age. An agent partnering with a senior must be sensitive to slower information processing and less multitasking~\cite{black2002elderly}.The agent must adapt its speaking rate to that of the senior user. It should also provide them with only the amount of information that they can process at a given time. An agent communicating with seniors should be assessed according to its capability to demonstrate these qualities. It should also be assessed according to what the user got out of the interaction. Was the senior able to use the information to get where they intended to go? Did they enjoy their chat with the agent? Would they use it again? Older people own fewer smartphones and have less access to the internet. By using an agent that is connected to a phone number, seniors can call the agent to get information about, for example, how to get to a doctor’s office~\cite{mehri2019cmu}.

\subsection{Research Questions} 
Future intelligent agent research should be driven by the exploration of a broad range of applications. This will see the user communities expose gaps between current technology and the new research questions to be addressed.

Research in education and healthcare applications will generally involve the need to preserve user privacy, particularly because of longitudinal data storage. These concerns will extend to other applications as more agents become personalized. Some research should address the tension between preserving privacy and having access to user models and audio-visual signals that can improve agent performance. The need to preserve privacy also poses infrastructure challenges concerning data sharing, which is desirable for reproducible research. Longitudinal data use also poses challenges to current machine learning algorithms. Many of the present algorithms rely on access to static datasets based on a large number of users. This is not practical when long-term data is needed from individual users.

Many of the above applications can potentially involve multiple interacting parties, i.e. multiple users and/or multiple agents.  For example, in assistive care, an agent might communicate with the patient and the caretaker together as well as separately. In education, an agent that supports student collaboration would deal with multiple users at the same time. New research challenges will spring form these situations related to knowing who is interacting and what their roles are, tracking the user state for each user and learning dialog policies for users with different objectives. Evaluation methodologies will need to account for multiple objectives as well.

Applications that explore a variety of domains and new user communities will likely reveal new challenges associated with particular types of language use, such as domain jargon, student idioms and slang, and linguistic code-switching amongst multiple languages. In such cases, the standard language processing approach of relying on pre-trained word embeddings is less useful. Some user populations may pose challenges associated with speech technology. Speech recognition is known to work poorly for children and for users with strong accents. Intelligent agents built to serve speakers of low resource languages or dialects will also require special speech and language recognition and generation technology development. 

Finally, application platforms that involve a variety of user populations will drive research that enables agents to serve more users. Working with narrow populations to serve specialized needs may involve people who are more or less comfortable with an agent and require different interaction strategies. Depending on the application, a user may assume different relationships with the agent, i.e. treating the agent as servant or tool that extends their own capabilities vs. a teacher or a facilitator. These roles will necessarily require different interaction strategies, both in actions and language use. For applications where an agent interacts with a very broad user population, there is the potential for interacting with people who have different styles and agendas, including adversarial users. Agents will need a flexible and robust dialog policy to handle a wide range of user types. In addition, systems will need strategies for learning about and adapting to user preferences/style both for understanding user intent and dealing with miscommunication. 

\subsection{Interdisciplinarity}
User-centered agents will exist in many forms: as a virtual assistant in wearables (e.g. smart watches, augmented reality glasses and hearables), in smart homes (e.g. smart speakers, TV sets) and as social robots (e.g. companions for the elderly, tutors for children and support for doctors). Development of multimodal systems, such as social robots, requires interdisciplinary research teams, since research projects may involve speech processing, natural language processing, computer vision, robotics, interaction design, cognitive science, and more. If intelligent agents are to be used for real end users, the actual content of the interaction is of paramount importance. True societal impact of the research requires long-term collaboration between the research community and the society they live in. Application-inspired research where intelligent agents are evaluated in healthcare and educational settings will add knowledge both in fundamental research and in specific domain areas. In order to achieve this, project teams need to be even more interdisciplinary by including both researchers in other fields (psychology, education science, medical science) and practitioners in schools and hospitals.

The evaluation of traditional task-oriented dialog systems has focused on objective measures of the quality of system components, task success, and user satisfaction as measured in post-use questionnaires~\cite{walker1997paradise}. In slot filling tasks, user satisfaction is often negatively correlated with a long time and a lot of effort to fulfill the task. In other domains, such as companions for seniors, this may not be the case. In the case of a health companion reminding a user to take medicine or contact friends and family, the user may also judge the quality of the interaction on other aspects (voice and speaking style, interesting content, level of engagement or adaptation). When developing companions for seniors, teams should include social and cognitive psychologists and practitioners from eldercare.

In other domains, such as education, user satisfaction during the interaction may be less appropriate as an evaluation criterion than the learning outcome. This calls for educators to be part of the team during the development and evaluation of educational intelligent agents. Similarly, intelligent agents that are support tools in healthcare require the involvement of medical researchers and hospital practitioners. Linguists would be needed for teams that develop applications with particular types of language use, such as kids’ jargon, air traffic control talk, or linguistic code-switching between multiple languages. Agents that target low resource languages will require special speech and language recognition and generation technology to be developed. This requires team members with additional language expertise and familiarity with the specific culture.

\subsection{Recommendations} 
Going forward, there should be a stronger focus on novel applications of intelligent agents, especially those that have not been well-covered in user-oriented agent research.

Research in user-oriented agents should uncover and address issues unique to these application areas. These applications should be:
\begin{itemize}
    \item generalizable
    \item dependant on the user pool
    \item sensitive to aspects of the individual user (e.g. their satisfaction, their cognitive or behavioral change, their speaking proficiency)
\end{itemize}

User-oriented intelligent agent research teams need to be interdisciplinary to tackle technological, scientific, and ethical/societal challenges.

\section{Infrastructure}

\subsection{Introduction}

With the advent of deep neural networks, a significant majority of research in AI concentrates on incrementally improving the performance of evaluation metrics on specific tasks and corpora. Thus the creation of tasks, corpora and evaluation metrics has a central role in defining the community's research direction. Solid infrastructure will facilitate effective intelligent agent research, as defined throughout this report. The sections below outline specific recommendations for (1) corpora, (2) tasks, (3) evaluation metrics and (4) tools. Creating solid infrastructure will ensure that research, even incremental advances, addresses the long-term challenges of intelligent agents.

\subsection{Corpora}
For the past 30 years the central role that data plays in Natural Language Processing (NLP) has been recognized. The creation and availability of dialog corpora has exponentially increased during that time. In fact, the number of available dialog corpora (speech-to-speech, text-to-text, etc) is so large that no one list of existing corpora can be exhaustive. 
A fairly comprehensive overview of many  (spoken) dialogue corpora can be found in \citet{serban2018survey}. There are several types of dialog corpora:

\begin{itemize}
    \item open-domain chit-chat conversations: SWITCHBOARD \citep{godfrey1992switchboard}, CALLHOME \citep{canavan1997callhome}, OpenSubtitles \citep{tiedemann2012parallel}, DailyDialog \citep{li2017dailydialog}, PersonaChat \citep{zhang2018personalizing}, Topical-Chat \citep{gopalakrishnan2019topical}; 
    \item information seeking dialogs: Let's Go (bus scheduling)\citep{raux2005let}, VERBMOBIL (meetings) \citep{wahlster2013verbmobil}, ATIS (travel reservations) \citep{dahl1994atis}, COMMUNICATOR \citep{bennett2002communicator},  Cambridge restaurant \citep{henderson2014dstc2,henderson2014dstc3}, MultiWOZ (complex travel planning tasks)  \citep{budzianowski2018multiwoz};
    \item question-answering: QuAC \cite{choi2018quac}, CoQA \citep{reddy2019coqa};
    \item  social media e.g. Yahoo news comments \citep{napoles2017}, Internet argument corpus \citep{abbott2016}, Reddit comments \citep{henderson2019repository};
  % \item visually grounded dialog: VisDial \citep{das2017visual}, CLEVR-Dialog \citep{kottur2019clevr};
\end{itemize}

Despite the ever-increasing number of dialog corpora, there are several key issues for future datasets that are identified and enumerated below. \\

\noindent
\textbf{4.2.1 Situated Corpora:} There is a dearth of ecologically valid, situated language interaction corpora: most of the available corpora focus on tasks that, even when ultimately involving entities in the real world (e.g. reserving a table at a restaurant), do not require the user to refer to entities that are physically, or at least virtually, present in the conversation. This is not to say that none of the earlier corpora referred to  physically or virtually present entities: for example, in expert-apprentice dialogues for assembling a toy pump from \citet{oviatt1992spoken}, the participants refer to a physical object; in both MapTask \citep{carletta97} and TRAINS \citep{trains95}, the participants refer to an external map (in MapTask, the two maps are inconsistent). \\

\noindent
\textbf{4.2.2 Multimodal Corpora: }Most available corpora do not include modalities beyond language, whether spoken or typed. This is beginning to change with datasets like VisDial \citep{das2017visual} and CLEVR-Dialog \citep{kottur2019clevr}. Nonetheless, to the best of our knowledge, there is no corpus with a variety of modalities that pertain to interactive agents (e.g., language, vision, speech, haptics, facial expressions, motion, prosody, proxemics, etc.). \\

\noindent
\textbf{4.2.3 Real vs Paid Users: }Many corpora are created with paid users and are therefore not reflective of legitimate conversations with real users \cite{ai2007comparing}. Past research has shown that paid users will complete only enough work to finish the task, while real users have an extrinsic motivation for interacting with an intelligent agent and will take more time to get what they want. While it may be difficult to collect certain corpora with real users, real users could be used in some capacity -- for example to understand common interaction patterns and motivate the instructions for paid users (e.g., if real users tend to switch a slot in the middle of a conversation, paid users should be instructed to do the same).\\

\noindent
\textbf{4.2.4 Research-Driven Corpora: }
Many corpora are \textit{application-driven} rather than \textit{research-driven}. New datasets should not necessarily be created just to build systems for a new application (e.g., booking a taxi). Rather they should facilitate research in a specific direction (e.g., task transfer, controllable systems). With the availability of corpora driving the majority of research, the collection of corpora can be thought of as the definition of a research challenge. For example, the structured intermediate labels of belief states and dialog acts in the multi-domain MultiWOZ corpus \citep{budzianowski2018multiwoz} resulted in research into domain generalizability \citep{mehri2019structured} and controlled generation \citep{chen-etal-2019-semantically}. Similarly, the Empathetic-Dialogues corpus allowed for research into controllable dialog generation \citep{lin2019moel}. Dialog corpora should collected with the underlying goal of understanding of what is needed for a particular research direction. This may initially necessitate the collection of diagnostic datasets that are limited in scope all the while allowing for targeted research in a specific direction.

\subsection{Shared Tasks}

Shared tasks or challenges are a proven mechanism for creating and encouraging specific research directions. While the creation of research-driven corpora may motivate research in a specific direction, shared tasks are necessary to facilitate the research beyond datasets. They provide both the motivation (competition, comparison of results with colleagues) and the means to achieve the research (tools and leaderboards, chat communication between participants, etc).

The infrastructure needed to conduct these shared tasks is important to the community. It facilitates participation and enables effective and comparable research in specific research directions. 

DARPA Communicator \citep{walker2001darpa} was the first instance of a shared task within this research community. It relied on full interactive evaluation, had a growth trajectory from a single task to multiple tasks, and was multimodal. There have been many similar shared tasks since then, including the annual DSTC competitions \citep{henderson2014dstc2,henderson2014dstc3}, the ConvAI competitions \cite{dinan2019second} and the Alexa Prize challenge \citep{ram2018conversational}. These shared tasks attract many participants, and can serve to highlight the shortcomings of intelligent agents. For example, the ConvAI2 challenge \citep{dinan2019second} revealed the weakness of automatic metrics at predicting the user impression of an agent. 

One important future direction for shared tasks is to conduct end-to-end, interactive assessment of the intelligent agent. This gains relevance since, increasingly, the majority of current research focuses on tasks such as response generation. Interactive assessment requires research that moves beyond a dataset and favors the shared task approach. The DialPort Portal \cite{lee2017dialport} is an example of a tool endowed with the ability to facilitate shared tasks by providing an interactive, user-centric environment. 
Many of the research directions highlighted throughout this report are potential long-term themes for shared tasks, including situated interaction, complex planning, multimodal dialog, and learning with limited data. Shared tasks will provide a common infrastructure and evaluation platform for these research directions, ensuring consistency during comparison and reproducibility.

\subsection{Evaluation}

Intelligent agent evaluation serves two purposes: a comparison between systems (or system variants) and an internal optimisation, (e.g., to be used as an objective during training). Just as the creation of corpora drives intelligent agent research, evaluation metrics define \textit{improvement} criteria and therefore strongly influence research. For example, an evaluation framework that emphasizes user satisfaction will result in research that promotes user-centric agents. As such, it is important to develop and promote evaluation metrics and frameworks that facilitate important directions of research (e.g., user-centric metrics to promote research into user-centric agents). 
%In this case, automatically computable evaluation metrics as well as direct user feedback have been used.

Standard automatic metrics for language generation include BLEU \citep{papineni2002bleu}, ROUGE \citep{lin2004rouge}, and METEOR \citep{denkowski2014meteor}. Recently, embedding-based metrics have been proposed include Greedy Matching \citep{rus2012comparison} and Skip-Thought cosine similarity \citep{kiros2015skip}. These metrics have been shown to be ineffective for dialog evaluation \citep{deriu2019survey,liu2016not}. For example, dialog-agent interaction is a very complex task with an infinite number of possible branches, (i.e., what to say, in which way, at which point in the dialog). Each dialog context has several valid responses, known as the one-to-many problem of dialog \citep{zhao2017learning}. Also, many current evaluation metrics rely on comparison with a ground truth, reference response which becomes ineffective when there are many potentially valid responses. This issue has started to be addressed through the use of multiple reference responses, through automatic retrieval \citep{galley2015deltableu,Sordoni2015ANN} and through data collection \citep{gupta2019investigating}. 

\citet{lowe2017towards} trained a model on quality annotations and attained strong correlations with human judgement. \citet{li2017adversarial} trained a model to predict whether a response was written by a human or a machine, and found that using this as a training objective resulted in improved performance. There is room for improvement, particularly in producing a general metric that: (1) does not suffer from the one-to-many problem of standard metrics that compare to a reference, (2) can generalize to many datasets and does not require explicit annotations specific to a particular dataset, (3) can correlate well with human judgement on a wide variety of tasks. 

Since intelligent agents are interactive, metrics must function effectively in interactive environments. Research in evaluation has largely focused on the task of response generation (i.e., given a fixed context, evaluate a generated response). While this task has proven challenging, it is unclear whether intelligent agents that perform well on the task of response generation will generalize to an interactive environment with a real user. In this interactive setting, it is necessary to assess agents with better automatic metrics and improved human evaluation. 

DialPort \citep{lee2017dialport} is a platform for interactive assessment of intelligent agents. Prior experiments on DialPort relied on feedback from the user (e.g., ratings). \citet{eskenazi2019beyond} found that leveraging the \textit{next user utterance} following the system response being assessed improves correlation with human judgement. This suggests that that interactive evaluation metrics can leverage additional context to improve quality assessment. Designing and incorporating an automatic metric for quality assessment into an interactive platform is a first step toward user-centric intelligent agent evaluation. 
\subsection{Tools and Platforms}

The availability of platforms, tools and baseline systems has a significant effect on the research that the intelligent agent community carries out. Once LSTMs  \citep{hochreiter1997long} became readily available in deep learning toolkits, they were widely used for research. Similarly, a thorough, well-documented repository of modern tools for building dialog systems would reduce the barrier for newcomers and ensure that they can compare their findings to strong baseline systems. 

Likewise, the creation of solid platforms for conducting research can reduce the entry barrier, and maintain the consistency and reproducibility of research. The OpenAI gym \citep{brockman2016openai} was created as a toolkit and platform for developing and assessing reinforcement learning algorithms. It has exhibited significant value both as an educational tool and as a common platform for research.

There are several tool repositories that exist for dialog, including the DialPort Tool Repository\footnote{http://dialport.ict.usc.edu/}, Facebook's ParlAI \citep{miller2017parlai}, Uber's Plato Research Dialog System\footnote{https://github.com/uber-research/plato-research-dialogue-system}, and RASA \citep{bocklisch2017rasa}. These repositories contain a collection of tools for building and deploying dialog systems. These tool repositories are a valuable resource to the community. Going forward, they need to gain in breadth (need to go beyond focused goals: e.g., industry applications) and include much more documentation (due to massive breadth). It is important to continue the development of a thorough repository of tools with strong documentation. Furthermore, it is important to have open source baseline intelligent agents that are thoroughly documented and evaluated. Such systems can serve as baselines for research, especially for newcomers to the field.

There is a shortage of shared platforms, similar to the OpenAI gym \cite{brockman2016openai}, for intelligent agent research. DialPort \citep{lee2017dialport} is a platform with a focus on interactive assessment. In addition to their tools, ParlAI \citep{miller2017parlai} allows for deployment and sharing of dialog systems. A thorough platform should be developed (and/or the present platforms be extended) that can support a breadth of research in intelligent agents, from baseline systems, to creation, assessment and deployment. Platforms should support multimodal and situated agents, knowledge graph integration and longitudinal user tracking. Additionally, this infrastructure would facilitate shared tasks in the domain of intelligent agents.

Tools and platforms should eventually be integrated through the concept of "plug and play", wherein focused search could be carried out on a specific component of a system without modification or re-implementation of the entire system. There are several variants of plug and play, allowing for flexibility of usage and therefore generality of the infrastructure. Researchers should be able to use infrastructure as a black box; for example, building an improved text-to-speech component for intelligent agents and assessing it with an intelligent agent. The infrastructure should also function in a translucent mode (i.e., some parameters of existing systems being easily modified) and transparent (i.e., significant changes can be made to existing components without complete re-implementation).

\subsection{Recommendations}

Infrastructure choices define and drive intelligent agent research directions. Corpora and shared tasks define research challenges for the community. Evaluation metrics determine what is judged to be an \textit{improvement}, and therefore what gets published. Tools can provide baseline systems, ensuring that research compares against strong models. There should be documentation for newcomers. Recommendations for each of these areas are provided below:

\begin{itemize}
    \item \textbf{Corpora: }The collection of situated and multimodal corpora is an important future direction. Also, dialog corpora should incorporate real users in some capacity. Finally, corpora should be research-driven rather than application-driven: a corpus should be collected to facilitate a specific direction of research (e.g., transfer learning, controllable systems, flexible dialog) rather than for a specific application.
    \item \textbf{Shared Tasks: }Shared tasks will require the development of a common infrastructure to promote effective and comparable research. Examples of potential areas for shared tasks are interactive assessment (for which the DialPort Portal, for example, can be used as a platform), situated interaction, complex planning, multimodal dialog and learning with limited data.
    \item \textbf{Evaluation: }Existing metrics (e.g., BLEU, METEOR) compute word-overlap and are insufficient for evaluating intelligent agents due to the one-to-many nature of their utterances. Therefore metrics need to be designed that account for these properties. Furthermore, since intelligent agents are inherently interactive, evaluation should occur in an interactive environment rather than on static tasks like response generation. Automatic metrics should be designed to effectively quantify the performance of systems for both response generation and in interactive assessment.
    \item \textbf{Tools and Platforms: }A common research platform should be developed or extended for intelligent agent research. This platform should be plug and play, support multimodal and situated interaction, and support research from baseline systems to creation, assessment and deployment. The platform should consist of a thorough, well-documented repository of tools for the development of intelligent agents. Such a platform would be long-lasting infrastructure that is a valuable resource for research, education and shared tasks.
\end{itemize}

\section{Dynamic Views of User-Agent Interaction}

One of the greatest challenges for intelligent agents is variability within and across users.  Users have different needs, preferred interaction styles, and comfort levels with agents. For social and companion systems, it can be important to consider the user’s opinions, age, and cultural background. Other factors may be important in other applications (e.g. student skill or knowledge level in educational systems). In addition, automatic speech recognition (ASR) performs poorly for some groups of users (e.g. children, the elderly, non-native speakers), which impacts the agents' efficacy. In order for agents to server a broad spectrum of users in a range of applications, systems will need the capability of learning about users.  Agents also need to be sensitive to the state of the user within a dialog, because of miscommunications, changing user interests and skills, etc.  Finally, as dialog technology evolves to provide new and improved capabilities, users will co-evolve with them, as evidenced by the dramatically different way people interact with virtual assistants now than in the demonstration dialog systems of the 90’s, i.e. users are more conversational (including more disfluent), and less uncertain. 

Current intelligent agents have limited capabilities for adapting to users. Virtual assistants that associate a user with a device can track user behaviors, such as website clicks, purchasing records, music choices, and may have access to the user’s calendar, which is useful for providing personalized services.  Regular users of smart speakers can be identified by voice to provide similar personalization.  However, these agents make little use of language or other modalities to assess the dynamic needs of a known user or respond adaptively to a new user. 
As described in this section, there have been initial research efforts in this direction aimed at recognizing user affective and cognitive states, as well as dynamic user modeling. More work will be needed, but in addition, user-centric dialog systems will need to handle the dynamics of the user-system, which will evolve as capabilities and roles change.

\subsection{Recognizing engagement, emotion and affect}
User-agent interaction can be significantly improved by recognizing the cognitive and affective state of the users. During human interaction, our emotions play an important role in our decision-making process. Our affective state is externalized through multiple communicative channels including facial expression, speech, language and body posture.  This information is perceived by interlocutors, who make perceptual judgements that influence their reactions toward us. Likewise, we can easily infer the cognitive state of others, realizing when they are paying attention, mind wandering or immersed in thoughts. For a user-oriented agent to be able to mimic these human-like responses, it is imperative to design algorithms that can recognize cues toward modeling the cognitive and emotional state of the user. 

Emotion recognition is one of the main challenges in affective computing~\cite{picard2000affective}. While there are several modalities that have been used to recognize emotions, the most common methods consider facial expression, speech or lexicon content. These modalities can be used in user-oriented agents depending on the context of the interaction. For example, for telephone and voiced-based virtual assistant systems, speech is the only modality available. Therefore, it is important to design speech-based algorithms that are robust against different channel, speaker, and environment variabilities. As labeled data with emotional descriptors are expensive to collect, it is important to leverage unsupervised approaches that can mitigate the mismatch between train and test conditions by using unlabeled data~\cite{abdelwahab2018domain,parthasarathy2019semi}. As ASR technology improves, automatic transcriptions can also be used for sentiment analysis. New advances in deep learning have created powerful word embeddings that have led to significant improvements in text-based sentiment analysis~\cite{dos2014deep,severyn2015twitter}. The advances have been triggered, in many cases, by the interest in understanding emotion in social media. This technology can also be incorporated in emotion-aware, user-oriented agents. Emotions can be recognized as well from facial expressions in scenarios where placing cameras is feasible (e.g., human robot interaction, and virtual kiosk). Important advances have been made on still images displaying clear facial expressions. However, there are still important challenges in face analysis from videos, especially when the subject is speaking~\cite{mariooryad2015facial,kim2014say}. The articulatory movements associated with speech cause changes in the facial appearance, introducing nuances that mask and regulate emotional cues. Furthermore, static representations from isolated frames may not provide an adequate representation of the emotion expressed in a video, as contextual information is not considered~\cite{ambadar2005deciphering}. Future advances in this area will require better models to infer emotional state of the user from videos, capturing the temporal evolution of the interplay between articulatory and emotional facial displays. 

Another important aspect of user-oriented agents is to estimate the engagement level of the user~\cite{castellano2009detecting,peters2009exploration}. This capability is particularly important for sustained interactions where the user can easily become distracted or disengaged. Any strategy to reengage the user should start by detecting signs of disengagement. Other important social signals that are relevant for a user-oriented agent to detect are cognitive load, turn taking, frustrations, politeness and disagreement. Advances in behavioral signal processing offer opportunity to infer these cues from multimodal sensors, improving the characterization of the user in the interaction~\cite{narayanan2013behavioral,vinciarelli2009social}. New developments in this area will facilitate the implementation of more intentional human-like strategies that aim to increase the engagement of the users, leading to better systems. 

\subsection{User Modeling}
Accounting for author/speaker variations has been shown to be useful in many NLP tasks, including sentiment analysis~\cite{volkova2013exploring}, comment recommendation~\cite{agarwal2011personalized}, and machine translation~\cite{mirkin2015motivating}. In conversational systems, user modeling has a long history~\cite{zukerman2001natural}. Systems have focused on different aspects of users, e.g., the expertise level of the user in a specific domain~\cite{hovy1987generating}, the user's intent and plan~\cite{moore1992exploiting} and the user's personality~\cite{devault_simsensei_2014,fung2016zara}. The Alexa Prize challenge inspired several efforts to leverage user information in the dialog policy, e.g.~\cite{bowden2019slugbot}. User modeling has also been employed for personalized topic suggestion in Alexa Prize socialbots, using a pre-defined mapping between personality types and topics~\cite{fang2017sounding}, a conditional random field sequence model with hand-crafted user and context features (\cite{ahmadvand2018emory}), or a weighted combination of latent modes with weights computed based on a neural representation of the utterance~\cite{cheng2019dynamic}. Modeling speakers with continuous embeddings for neural conversation models is also studied in~\cite{li2016persona}. Directly or indirectly, recognizing emotion, engagement and other social signals can contribute to understanding a user more globally. 

\subsection{Recognizing user knowledge, mental states, theory-of-mind}

The complexity and sophistication of (spoken) language tends to be masked by the apparent ease with which we, as human beings, use it.  As a consequence, engineered solutions are often dominated by a somewhat naive perspective involving the coding and decoding of messages passing from one brain (the sender) to another brain (the receiver). In reality, language is better viewed as an emergent property of the dynamic coupling between cognitive units that serves to facilitate distributed sense-making through cooperative behaviors.  Furthermore, the contemporary view is that language is based on the co-evolution of two key traits - ostensive-inferential communication and recursive mind-reading - including ‘Theory-of-Mind’~\cite{scott2014speaking}.

These perspectives on language not only place strong emphasis on the importance of pragmatics~\cite{bar2017communicative}, but they are also founded on an implicit assumption that interlocutors are conspecifics, and hence share significant priors.  Indeed, evidence suggests that people draw on representations of their own abilities expressed as predictive models in order to interpret the behaviours of others, and this is thought to be a key enabler for efficient recursive mind-reading and hence for language-based interaction.

So, if (spoken) language interaction between human beings is grounded through shared experiences, representations and priors, to what extent is it possible to construct a technology that is intended to replace one of the participants?  There will be an inevitable mismatch between the capabilities of the two partners, and coupled ostensive recursive mind-reading (i.e. full language) cannot emerge.  Hence there may be a fundamental limit to the language-based interaction that can take place between humans and artificial agents~\cite{moore2017spoken}.  These are issues that need to be investigated.

\subsection{Agent Presentation}
The user may be highly influenced by the way the agent presents itself, in terms of willingness to interact, style of interaction, and evaluation of the interaction. The agents’ self-presentation includes aspects of identity, such as name, backstory and job/role, language usage, including word choice, utterance structure, and linguistic register, and, in multi-modal settings, the voice and physical appearance. Some common goals for agent-presentation include:
\begin{itemize}
    \item One that is familiar to the user, to set expectations for the interaction based on the familiar associations
    \item One that will engage the user and make interaction more attractive
    \item One that will inspire trust in the information conveyed by the agent
    \item One that will set up appropriate affordances to make clear what the agent can or cannot do (well).
\end{itemize}

Some of these goals may be in conflict with each other. E.g., a highly familiar and/or engaging personality might attract the user initially but might influence the user to assume that the system is more capable and able to handle more complex interaction than it actually can.
Another conflict is whether the agent should follow a very simple protocol (always using the same language construction for the same function) or have more varied interaction. The former may be easier for users to learn and adapt to, but will be more boring and perhaps off putting in its redundancy. The latter will be more engaging and human-like, but may lead to more problems if users assume that they can say anything to an agent with only limited understanding abilities. It may be desirable to adapt the agent presentation to the desires of the user as to their ideal interlocutor (e.g. selection of voice, register, gender, age, human-ness, appearance, clothing,...)

One important issue concerns the role of language in human-agent interaction.  For example, studies into the usage of smart assistants suggest that, far from engaging in a promised natural `conversational' interaction, users tend to resort to formulaic language and focus on a handful of niche applications which work for them~\cite{moore2017spoken}.  Given the pace of technological development, it might be expected that the capabilities of such devices will improve steadily, but according to~\newcite{phillips2006applications} there is a ‘habitability gap’ in which usability drops as flexibility increases.

It has been hypothesized that the habitability gap is a manifestation of the ‘uncanny valley’ effect whereby a near human-looking artefact (such as a humanoid robot) can trigger feelings of eeriness and repulsion~\cite{mori2012uncanny}.  In particular, a Bayesian model of the uncanny valley effect~\cite{moore2012bayesian} reveals that it can be caused by misaligned perceptual cues.  Hence, a device with an inappropriate voice can create unnecessary confusion in a user.  For example, the use of human-like voices for artificial devices encourages users to overestimate their linguistic and cognitive capabilities.

The Bayesian model of the uncanny valley effect suggests that the habitability gap can only be avoided if the visual, vocal, behavioral and cognitive affordances of an artefact are aligned.  Given that the state-of-the-art in these areas varies significantly, this means that the capabilities of an artificial agent should be determined by the affordance with the lowest capability (Moore, 2017; Wilson and Moore, 2017.  In other words, emulating a human is a recipe for failure, rather “it is better to be a good machine than a bad person”~\cite{balentine2007s}.

\subsection{Agent-User Relationship}
It is useful for the system to adapt to different types of users.  At the same time, users will be adapting to their understanding of how the system works and what seems to work well (even sometimes inappropriately, e.g. hyperarticulation of words, which is not well covered in a speech recognizer’s acoustic model), or they may be changing their needs and use of the system as they learn more about its capabilities. In particular, a user who is interacting  with the system for the first time may need a lot of guidance in terms of explanations, examples, enumerated choices at various stages. More expert users will not need as much of this kind of help and may find it annoying and wasting time. A user-focused system should adapt to the level of expertise of the user, which includes being able to use and understand language that the user is comfortable with, providing explanations to novices and short-cuts to experts. Entraining  to the user (using the same words, prosodic patterns or gestures, in an agent) may be part of the strategy to achieve this. For example, in the context of educational spoken dialog, a robot that both entrained to and spoke socially with students resulted in significantly more student learning~\cite{lubold2018automated}.   Converseley, dialog systems have also taken advantage of user entrainment to the system by making lexical, syntactic, and prosodic system utterance choices believed to be more conducive to system success~\cite{stoyanchev2009lexical,fandrianto2012prosodic,lopes2013automated}. It may also be useful for the agent to use strategies to encourage other types of change in user behavior, though ethical issues will arise here. Intentional actions such as these must be carefully assessed for their utility. The agent should also be adaptive to the level of communication problems; e.g. in a noisy environment, it may be helpful to have shorter utterances and more explicit feedback of what was understood, and/or giving only summary information rather than more elaborated context. Incremental understanding and generation are likely to important for the more advanced turn-taking necessary for effective communication.

The agent-user relationship becomes much more challenging for scenarios that involve multiple agents and/or users.  When in a group setting, should the agent address the group or an individual, and which user model(s) should be active? In contexts with changing group membership, the same person can play different roles or have different expertise as the groups change, which impacts the user model for that person. Just as behavior of an individual user can be affected by emotion and background knowledge, the interactions of people in a group are affected by group social dynamics and their knowledge of the expertise of each other.  How do the agents contribute to group dynamics, and to what extent can/should agents aim to change group dynamics?  For example, when pairs of students working together were dynamically supported by a dialog agent, student learning improved~\cite{kumar2007tutorial}. When there are multiple agents, how do they coordinate their communication and efforts to build user models?

\subsection{Recommendations}
The challenges with user variability that current systems face and the potential for future multi-party applications suggest the following areas will be important for future research:
\begin{itemize}
    \item Move research from generic, static views of the user, the agent and their relationship to personalized and dynamic models; and
    \item Move from human-computer to multi-party, where the agent learns about different humans and understands it specialized expertise relative to other agents
In order to ensure that research is addressing problems relevant to these issues, it is important that research emphasize scenarios that involve longer conversations (beyond 1-2 turns) and/or more frequent interactions from the same user or groups of users.
\end{itemize}

\section{Building Dialogue Systems in Low-Resource Conditions}

\subsection{Introduction}
Current commercial technologies for natural language processing have come to rely predominantly on supervised learning. This requires impressive quantities of gold standard, annotated data for training, tuning, and evaluation. However, there are many scenarios and applications where there simply is not enough annotated data for system construction when using supervised learning. Consider, for example, multimodal input data streams that can be collected by an agent that communicates with human users. It is equipped for both real-time capture and generation of audio, video, and motion data. Datasets recorded during interactive sessions may vary in quantity and quality by channel, and they may not be synchronized or sampled at similar rates across the channels. This variation complicates the conversion of data into shared representations for downstream cross-modal alignment, annotation, and analysis (see a detailed discussion in Section~\ref{sec:mml}). Furthermore, even if the multimodal data that has been collected is abundant, the annotated portion may not be sufficient for supervised learning in system construction. Annotated data may be limited due to the time and cost of ground-truthing multimodal data for new domains, new tasks, or new languages. And the limited availability of multimodal data, that contains identifiable information (face, voice, language spoken) may be due to privacy issues (see detailed discussions in Section~\ref{sec:ethics}). To arrive at a definition of future research directions, this section first examines current state-of-the-art approaches under different scenarios.

\subsection{Current Approaches}
The conditions of resources can be categorized along along two dimensions:  1) the amount of in-domain data and 2) the amount of annotated data. Table~\ref{tbl:less} shows the possible combinations.

\begin{table}[h!]
\centering
\begin{tabular}{l|lll}  \hline
                                            & No Data & Limited Data & Abundant Data   \\ \hline
No annotations                              & Case 1  & Case 2       & Case 4          \\ \hline
Limited annotations                      & X       & Case 3       & Case 5   \\ \hline
Abundant annotations                        & X       & X            & Case 6 ("Solved")  \\ \hline
\end{tabular}
\label{tbl:less}
\caption{Current approaches for in-domain and annotated data }

\end{table}

\subsubsection{Learning when there is no data (column 1)}
\textbf{Wizard-of-Oz}: Many intelligent agent projects, whether in academia or industry, begin with an idea, but no available relevant data (Column 1). To bootstrap these projects, researchers typically resort to proxy methods for data collection such as Wizard-of-Oz (WoZ) systems~\cite{dahlback1993wizard,okamoto2001wizard,budzianowski2018multiwoz}, where a human operator plays the role of the as-yet-non-existent system, interacting with a human user. These interfaces are challenging to build, involving critical decisions that are likely to impact the rest of the project. The key trade-offs faced by WoZ system designers concern the information available to the wizard and the amount of freedom given to a wizard to formulate their responses to the user. For instance, the wizard may have full access to direct audio and video of the user, ensuring accurate perception of the user’s actions. Alternately, the wizard may only see written transcripts of the user’s speech since output by the system’s ASR includes errors. This more accurately represents the type of input stream that an intelligent agent would need to handle. Other arrangements also exist: they could be allowed to speak or type freely without constraint on the content or form of their response, or they might be restricted to select actions or phrases from a limited predefined set presented to them, for example, as buttons on a screen. In general, the need to get a correct response to the user in as close to real time as possible will influence interface design choices.

Typically, the wizard or wizards’ knowledge is relied upon to manage the content and pacing of the dialog. The users interacting with the WoZ are asked to perform a given task communicating with WoZ interface, but are not informed that a human is “behind the curtain” performing the work of one or more of the system’s components (for example, dialogue management, speech recognition, speech generation, natural language understanding, natural language generation, reasoner). Thus the user together with the wizard generates the data that is needed to train the models for automating those components.  

Unfortunately, WoZ data collection is typically complex, time-consuming, and expensive~\cite{rieser2008learning}. If users in academic or government studies provide personally identifying information, there are additional requirements that apply to the experimental design protocols, as regulated by an institutional review board (IRB). Thus WoZ systems are generally limited in scale and often used only in very early stages of agent development~\cite{munteanu2000mdwoz}. Crowdsourcing has recently been used to address these limitations ~\cite{lasecki2013real,huang2018crowd}. This increases the scale of the WoZ paradigm toward large numbers of wizards (i.e. crowd workers) and thus users. 

\textbf{Rule-based Frameworks} Another approach to bootstrapping a research system when no training data is available is to construct a fully-automated system by relying on hand-crafted rules and manually-selected datasets to test those rules in the process of evaluating the full system~\cite{larsson2000information,bohus2009ravenclaw,zhao2016reinforest}. This approach offers the near-term benefits of being both more immediately scalable (though finding enough users may not be easy) and more amenable to data collection for machine learning since it relies on some amount of explainable machine representation of knowledge to perform its task. The main issues for such hand-crafted systems are that they are very costly to build. Developers must combine a large number of complex technologies. Yet they also permit a very wide range of user experiences. This makes the data too highly variable for the eventual training of machine learning models for system components. Often research teams do not collectively have the range of skills, resources, or time to develop and apply the methods needed for such iterative design to create operational, high quality systems.

\subsubsection{Learning from limited data (column 2)}
Machine learning methods, especially supervised deep learning models require a tremendous amount of training data to reach good performance. As a result, when only limited in-domain (either annotated or unannotated) data is available, the direct application of supervised learning falls short. Therefore, a rule-based framework is still one of the top choices when creating intelligent agents with only limited data. Existing research has explored the following approaches to empower an agent when faced with limited data. 

\textbf{Integrate rules into learning systems}: rules can be combined with statistics to improve the robustness of a dialog manager or natural language understanding module. \newcite{lison2016opendial} combines probabilistic rules with Bayesian learning in a dialog manager so that the probability that an if statement is triggered is learned from data rather than set by manual heuristics. Rules can also be applied to dialog state tracking~\cite{wang2013simple,sun2014generalized}. This significantly improves performance in low resource settings. Recently rules have shown promise when integrated with end-to-end dialog models~\cite{williams2017hybrid,razumovskaiaincorporating}, thus reducing the amount of training samples that are needed needed. 

\textbf{Transfer Learning}: another well-studied approach is to leverage data from other domains and enable knowledge transfer from models trained in a source domain to a target domain. Related research areas include transfer learning and zero (few)-shot learning. \newcite{chen2016zero} developed an intent classifier that can predict new intent labels that are not included in the training data. \newcite{bapna2017towards} extended the idea to the slot-filling module in order to track novel slot types. Both papers leverage a natural language description of the label (intent or slot-type) in order to learn a semantic embedding of the label space. Then, given any new labels, the model can still make predictions. Moreover, there has been extensive work on learning domain-adaptable dialog policy by first training a dialog policy on K previous domains, and then testing the policy on the K+1st new domain. \newcite{gasic2014gaussian} used the Gaussian Process with cross-domain kernel functions. The resulting policy can leverage the experience from other domains to make educated decisions in a new one. Finally, zero-shot learning has also been applied to natural language generation. \newcite{wen2016multi} used delexicalized data to synthetically generate NLG training data for a new domain. \newcite{zhao2018zero} proposed an action matching algorithm to transfer knowledge of an end-to-end generation-based dialog system. This allows an end-to-end system to operate in a completely new domain.

\subsubsection{Learning from limited annotated data (column 3)}
The third case (column 3) refers to when there is a vast amount of data in the same language or even the same domain, but there is not enough annotation of that data to train supervised models for natural language understanding or state tracking. For example, there is a large amount of human-human call center data, however, much of it is not annotated. In this setting, current state-of-the-art approach includes:

\textbf{Unsupervised end-to-end dialog system} uses end-to-end (E2E) dialog models that directly train a generative model on top of conversational data. Different from modular systems~\cite{zhao2016towards}, E2E dialog agents do not require annotation or system dialog acts for natural language understanding (NLU). E2E dialog systems can be divided into generative models~\cite{vinyals2015neural} and retrieval-based systems~\cite{lowe2015ubuntu}. However, despite promisng results, E2E systems as them today suffer from the following limitations. Significant amount effort has been invested to improve E2E dialog models. For example, generative models are known to generate dull rather than relevant responses. A number of solutions have been proposed such as maximum mutual-information training~\cite{li2015diversity}, latent variable models~\cite{zhao2017learning}, etc. A second limitation for E2E dialog systems is that they cannot easily interface with an external structured data source. This prohibits them from serving information from a database~\cite{zhao2016towards,dhingra2016towards,williams2017hybrid}. A third open-ended research question for E2E system is evaluation. Since there is no human annotation available, systems cannot be compared by using standard metrics, like accuracy. Most current methods compare the system response against human reference responses. Therefore, current mainstream metrics, e.g. BLEU~\cite{papineni2002bleu}, correlate poorly with actual system performance~\cite{liu2016not}.

\textbf{Representation learning \& Semi-supervised learning}
In case 5, the state-of-the-art methods used to create intelligent agents are unsupervised representation learning and semi-supervised learning. Examples of pre-trained language models include BERT~\cite{devlin2018bert} and RoBERTa~\cite{liu2019roberta}. First a language model is trained with self-supervised objectives, for example masking language models on massive amounts of unannotated data. Then the resulting representation is fine-tuned on a small set of labelled data. 

\subsection{Recommendations}
Currently, machine learning approaches for dialog agents still struggle in the low resource setting, yet there are many promising research venue that worth of exploring. We recommend investing in the following directions in order to enable practical framework for building future dialog agents.
\begin{itemize}
    \item A systematic way to introduce prior knowledge as an input to ML models.
    \item Community effort to make pretraining models well-documented, thoroughly tested and widely available.
    \item Open-source, robust and reusable tools for solved tasks, e.g. wizard-of-oz interface, data collection tools, human evaluation platform etc.
    \item A principled approach for system design, data collection, resources management, and evaluation for new AI projects with limited/no data
    \item New machine learning methods that can better utilize knowledge from other domains or unannotated data.
    \item Shared tasks that standardize the evaluation of dialog agents under low resources setting, and regular benchmark comparison across various methods.
\end{itemize}

\section{Multimodal, grounded and situated interaction} 
\label{sec:mml}
\subsection{Introduction} 
Human communication is, by nature, multimodal, where interlocutors make use of a combination of communicative systems (‘modalities’), each serving as physical ‘carriers’ of messages (Bunt, 1995). In situated human dialogues, several human senses contribute to the overall experience (sight, hearing, touch, and smell). In human-human interaction, the various available modalities may be communicated sequentially or in parallel, carrying information that may be redundant or complementary, and the content may be transmitted deliberately or unconsciously~\cite{turk2014multimodal}. The voice carries both verbal and non-verbal information, where the latter may be transmitted by a combination of voice quality and utterance melody that convey the speaker’s personality traits, affective state, and willingness to talk. The face also carries significant information in situated interaction: an active listener may use facial gestures, as well as vocal cues, while the speaker is talking, to indicate their engagement along with nods of their head, for example, so as to provide listening feedback with facial expressions that convey emotional state (such as level of certainty or understanding). Eye gaze is yet another source of communication during dialogue that is fundamental in social and situated interaction; listeners may rely on it to assess the speaker’s focus of attention or to regulate speaker shifts. In situated interactions, joint visual attention is the process by which speakers coordinate what they are attending to, ensuring that they focus on the same object of interest. Mutual gaze, in particular, occurs when interlocutors look at each other during face-to-face interactions. This type of eye gaze has been shown to differentiate roles in dialogue: it has been reported that people look at the other party in an interaction nearly twice as much (75\%) when listening, as when they are speaking (41\%)~\cite{vertegaal2001eye}.

Computers have been historically uni-modal, with the earliest computer input and output displays provided by teletype machines that were limited to the visual modality of text. With the advent of GUI-based computer interfaces, the range of modalities available to users was extended to accept their input by mouse and to generate output for them by graphical interfaces and audio channels.  The commercial success of smartphones in the last decade has led to rapid development of multimodal-multisensory interfaces that now combine both active input (speech, touch, typing) that users are aware of, with the less obvious, or possibly covert, input signals that built-in sensors detect and that users may not be aware of, such as gaze, location, acceleration, proximity, and tilt~\cite{oviatt2015paradigm}. Today, smartphones also include biometric sensors that can identify the user (facial recognition or 
 fingerprint readers) and new sensors that can track human movements (RGBD cameras for face tracking, radar for gesture recognition). In the future, user-oriented intelligent agents will, similarly, be constructed to be coupled to an individual (via smartphones or wearables, placed directly on an owner’s body) or to other locations (via smart speakers or social robots). For such intelligent agents to engage more extensively in situated interactions with dialogue partners, they will need to be equipped with a wide range of sensors that feed built-in analytic components that may seek to predict or infer the emotional states and physical actions of humans with which they are communicating.  

Truly user-centered agents will have to be built with communicative abilities that enable them to interact with people, and possibly also with other robots, in an intuitive and socially acceptable manner. Social signal processing is a growing research area that aims to endow machines with the ability to sense and interpret human social behaviors, such as turn regulation, emotion, personality, dominance and rapport~\cite{vinciarelli2009social}. The multimodal behavioral cues that these systems will need to interpret, in conjunction with the user’s utterances, include: vocal behavior (e.g. prosody and voice quality), face and eye behavior (e.g. facial expressions and gaze patterns), gesture and posture (e.g. hand gestures and posture), proxemics (e.g. distance and F-formation). 

If domestic social robots are to share physical space with humans, they will need to be able to navigate among people, understand people's physical actions, and understand the significance (to humans) of objects that occupy their shared, common environment. In collaborative interactions, human participants engage in constructing and maintaining a shared conception of a problem at hand~\cite{dillenbourg1996grounding}. They do this by constantly establishing, maintaining, and repairing common ground~\cite{clark1989contributing}. In this process of grounding, the listener relates to what has been said and conveys that they are following with questions, clarifications, and feedback.  Sub-tasks within this (common) grounding process involve joint attention, engagement and turn-taking. Together,  their verbal and non-verbal feedback communicate their willingness and ability to (a) continue the interaction, (b) perceive the message, (c) understand the message, and (d) accept the message, including signaling attitudinal reactions.  

In situated interaction and robotics in particular, there is a distinct, relevant concept referred to as symbol grounding~\cite{coradeschi2013short}. A special case of this is perceptual anchoring, which is the process of connecting higher level symbolic information, such as words, to physical objects or their sensor data~\cite{harnad1990symbol}.  Collaborative robots will need to be able to engage concurrently in both types of grounding, that is, maintaining conversational common ground with interlocutors and performing perceptual anchoring in their environment.  Robots that simulate joint attention mechanisms have been widely shown to help disambiguate spatial references by making object referral appear more natural, pleasant and efficient~\cite{mehlmann2014exploring}. Robots involved in tasks where they engage with objects or their environment, will need to, at appropriate times, switch their gaze to their conversational partners. This coordination is essential for an appropriate perception of mutual gaze, but it is also an essential step for guiding the conversational partner's attention and establishing the common focus of joint attention. The simple establishment of meaningful and timely mutual gaze can lead to extended interactions and an increase in human attention when interacting with robots~\cite{ito2004robots}. 

\subsection{Research Themes} 
\subsubsection{Levels of representation / abstraction / granularity / alignment / incrementality}
\begin{figure}[!h]
 \label{fig:mml}
    \centering
        \includegraphics[width=0.8\textwidth]{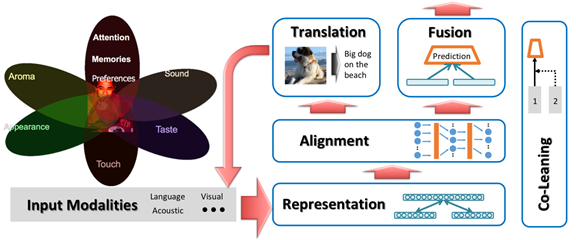} 
        \caption{Taxonomy for multimodal learning.}
\end{figure}

An important research endeavor is to develop foundational methods to analyze, model, and represent multimodal information, whether it is the input to, or the output of, user-oriented agents. These methods are characterized as foundational since they underlie the technical framework for information encoding and decoding that is essential to many applications and research topics. The unique and challenging aspect of multimodal research is the heterogeneity of multimodal sources of data and the challenges in integrating and interpreting such heterogeneous data coming from multiple modalities. Each modality has its own characteristics that have to be considered~\cite{maragos2008multimodal}. For example, images are 2-dimensional (or sometimes even 3-dimensional), while speech signals are 1-dimensional. The way we describe the visual appearance of a bird or another animal will be quite different from the way we describe the sounds they produce. Creating  computer systems that are able to make sense of such multimodal data entails many fundamental problems that we group into five main classes, following the taxonomy of~\cite{baltruvsaitis2018multimodal}: representation, alignment, fusion, translation, and co-learning in Figure~\ref{fig:mml}. 

\textbf{Representation}: Learning how a computer can internally represent the heterogeneous data in multiple modalities. These computational representations should be designed for both efficient modeling and better visualization. For example, a joint representation that captures how a person both looks and sounds when they are happy will allow computer systems to better recognize human emotions than systems that treat such visual and acoustic modalities of data independently. These joint representations will be most efficient when they can take advantage of the naturally-occurring dependencies between modalities. Another design objective is to improve the interpretability of multimodal data. By identifying commonalities and differences across multimodal data, we seek multimodal representations that best bridge the gap between continuous versus discrete data, and between statistical versus symbolic data. 

\textbf{Alignment}: Establishing spatial and temporal connections between events across modalities. For example, when reading the caption of an image, alignment is the process where words are grounded to specific objects or groups of objects in the image. Other examples of alignment include automatic video capturing, and identifying the acoustic source in a video. One challenge in alignment is dealing with data streams that have been recorded at different sampling rates (e.g., continuous signals versus discrete events). Alignments may thus require defining a similarity metric between modalities to identify the connection points across the streams. The alignments may need to be anchored temporally, as when we align the audio track and the images in a video, or they may need to be anchored spatial, as when we try to morph between two face images.

\textbf{Fusion}: Combining information from two or more distinct sources to uncover or predict a pattern or trait of interest. Examples of multimodal fusion include multimodal emotion recognition, audio-visual speech recognition, and audio-visual speaker verification. The sources of information found across modalities may be redundant which helps increase robustness, or they may be complementary which often helps increase accuracy. The challenges are multiple since, a priori, it may not be known in advance which sources will need to be synchronized, or even whether they have the same sampling rate. For any source, some modalities may be incomplete with missing data, while some modalities may provide continuous streams of data, but other modalities may be intrinsically discrete in providing data for interpretation about given events. 
 
\textbf{Translation}: Transformation, mapping from one modality into another. Examples include speech-driven animation and text-based image retrieval. The foundational methods learn the relationship between streams of data, capturing their dependencies. The goal of translation can be generative in nature, creating a new instance previously not observed in one modality, based on given information from another modality. The goal can also be for building descriptive models, where having learned inter-modality dependencies, one modality can be used to increase the characterization of another modality (e.g., describing the information conveyed in an image).  

\textbf{Co-learning}: Transferring knowledge from one modality to help with the predicting or modeling task in another modality. Co-learning examples are more technically complex in nature, building on established within-modality processing expertise. Co-learning aims to leverage the rich information available in one modality, for the learning of another modality, which may have only limited resources (e.g., small number of examples with limited annotations or noisy input). Examples of co-learning algorithms include co-training, zero-shot learning and concept learning. Recent examples include: the use of information extraction in text processing to guide entity detection in computer vision algorithms, such as for visual object recognition, when an implicit instrument argument can be inferred for an event mentioned in a caption accompanying an image~\cite{subburathinam2019cross} and multitask learning that exploits a text-guided semantic space to select the most relevant visual objects for novel visual activity recognition~\cite{eum2019object}. 

The challenges that arise in multimodal processing may involve various combinations of these five categories. Importantly, cross-cutting research in these areas will open opportunities to better understand and interpret multimodal data across domains, providing resources and tools for wider sharing across the interdisciplinary community of researchers. These tools may be generic, for working across a range of problems, or they may be specific to given problems or modalities. 

\subsubsection{Agent Input: Information Fusion}
Multimodal fusion is a core requirement in advanced user-oriented agents that are able to understand explicit and implicit information conveyed by users. Users display important information through various communication channels including their speech, facial expressions, hand gestures, and body posture. The information is not uniformly conveyed across time and modality. Studies have revealed a complex temporal interplay between communicative messages including lexical, semantic, emotional, and idiosyncratic traits~\cite{busso2007interrelation}. This interplay has also been observed across communicative channels, where less constrained modalities are used to convey communicative messages~\cite{busso2007joint}. This temporal and spatial interplay between communicative messages and channels has direct implications on how a user-oriented agent should interact with users, sensing, fusing and interpreting information obtained across multiple modalities.
 
From a machine learning perspective, multimodal fusion can be viewed as the process where features from different modalities are mapped into a common representation that is trained to maximize the performance of a classification or regression task~\cite{baltruvsaitis2018multimodal}. Recent advances in deep learning has resulted in powerful embeddings that encode numeric (e.g., speech, images) and symbolic (e.g., text) modalities. These embeddings can be systematically combined, leveraging the redundant and complementary information provided across modalities. We can obtain more robust models in this way, especially when the reliability of some modalities is affected, when some information is missing, or when a disagreement exists across a subset of the modalities. For example, the user’s face may be a more communicative channel to convey positive or negative emotional traits than their speech. On occasion however, occlusions or non-frontal views of the user may lead to unreliable or missing data. During those segments, using speech or lexical information alone may yield more robust predictions. This example shows that, when the data is available, leveraging redundant information conveyed across modalities can increase the performance of a system.
 
The challenge for a user-oriented agent is to interpret sensing information from the user coming from multiple modalities, taking into consideration the underlying complexity in the way humans interact and communicate with each other. The architecture of the models should be robust to partial information. For example, when a user is listening to the agent and extracting lexical information from the agent’s spoken message, the user will be most likely to limit their vocal responses to backchannel information. During those segments, their facial expressions may be more informative about their cognitive and emotional states than their vocal signals. By contrast, when the user is speaking, the information in their facial expressions will be less discriminative for tasks such as the recognition of emotion, mood or personality traits due to the speech articulatory movements that affect the facial appearance. Likewise, as multimodal databases with rich labels for relevant tasks are often limited, we should be able to train models with unimodal datasets that have partial information. Ideally, we should also build unsupervised formulations to leverage found rich multimodal data without labels. The architecture of the intelligent agent should also be able to assess the expertise of each modality, by learning from the data the appropriate weights to combine multimodal information. Advances in multimodal processing that address these challenges will facilitate improved user-oriented agents that are more aware of the user and can respond accordingly.

\subsection{Agent Output: Coordination of Multimodal Agent Behaviors}
Multimodal agents not only need to perceive multiple modalities, they  should also be able to actively transmit information across various modalities so as to express their own intentions and internal states to their human users. Such multimodal output generation presents both challenges and opportunities that fall into two broad areas: human-like multimodal rendering and context-aware modality combination.
 
The first area generally focuses on animating embodied conversational agents, whether virtual or robotic. Originating with early research on talking heads~\cite{cosatto2000photo,massaro2003computer}, this research also includes gestures and full-body animation of virtual agents~\cite{cassell2001more}. A large amount of this work focuses on realism and synchronization of several modalities. For example, \newcite{ng2010synchronized} present a model for generating and synchronizing hand gestures from arbitrary text to accompany a spoken rendering of the same text. More recently, \newcite{takeuchi2017speech} have investigated the use of deep learning techniques to predict hand gestures directly from the speech signal (MFCC), albeit with limited success. Finally, recent work (and media hype) focused on “deep fakes” showcases the application of modern deep learning technologies to multimodal rendering of humans~\cite{suwajanakorn2017synthesizing,chan2019everybody}. While such technology can now achieve near perfect photo-realism and multimodal synchronization, much less is known about the relationships between various modalities and concurrent conversational features, such as the speakers’ intent or inner state. Earlier work from~\newcite{cassell2001more} and \newcite{traum2008talking} has explored models of embodied conversations and their implications on multimodal rendering of virtual humans. More research on these topics would be greatly beneficial to our understanding of both human-human and human-machine communication.
 
The second area of multimodal output generation concerns the use of different modalities and affordances to accommodate different environments, situations, or user tasks and needs. For example, \newcite{costa2011adapting} describe a multimodal fission module that adapts the presentation of content through video, audio and haptic outputs for impaired users. \newcite{teixeira2011output} introduce a general architecture for adaptive multimodal fission and its application to a tele-rehabilitation service for the elderly. Both works highlight the potential of adaptive multimodal output generation to better accommodate a wider range of user populations, by incorporating design flexibility for more inclusive technology.

In addition, research is needed to better understand the effect of combining the generation of natural language and visual information on communication. One possible direction is to combine deep learning approaches to Natural Language Generation with image or multimodal representations (e.g. those used by image captioning work such as~\cite{xu2015show} to build general models that can generate multimodal output, based on the agent’s internal semantic representations.

\subsubsection{Incrementality}
When humans engage in face-to-face spoken interaction, listeners usually provide brief audio-visual feedback cues while their conversational partner is talking. Such listener feedback may occur concurrently in multiple forms, such as head nods, eyebrow movements, or facial expressions (visually) or conventionalized, vocal tokens of agreements (verbally).  By continuously providing feedback on what is being said, the active listener participates in the collaborative communicative process, helping establish and maintain common ground. Today’s state-of-the-art dialogue systems are neither receptive, nor able to change their output, while in the process of speaking, should the listener convey confusion, irritation, or other emotions as signals that should be addressed.  If we want to be able to build spoken dialogue systems that can achieve the responsiveness and flexibility found in human-human interaction, then it is essential that they process information incrementally rather than in utterance-sized chunks~\cite{dohsaka1997system,gervits2018pardon}. A conversational dialogue agent has to be able to decide: when to take a turn (controlling the floor as the next speaker), when to provide verbal feedback (without taking the floor), and when to remain silent. To do so, the agent must predict the end of the human speaker’s utterances, tracking the timing and pacing of turns in the conversation. The agent must also decide if its feedback should be verbal or non-verbal (head nod, eyebrow raise or smile). If it decides to produce verbal feedback, then it has to be able to decide if it should use a feedback token (such as “okay”, “mhm”) or if it should reprise fragments of what it has heard from the user to show positive evidence that it is, in effect, paying attention , echoing conversational content by way of a word or phrase that has been picked up. Finally, it has to select   the prosodic realization of the feedback token to convey the appropriate affective state~\cite{neiberg2013semi}. 

Multimodal systems for agents need to be developed according to principles such as “Inclusive Design” or “Design for All” to be accommodating in situations where users present with varying levels of physical or cognitive abilities. According to~\cite{obrenovic2007universal}, there are two sources of constraints in human-computer interaction: user- related constraints (such as user features, user preferences, and user state) and external constraints (such as device, environment, or social context). Furthermore these constraints may interact and change during an interaction. For example, the level of environmental  noise  may increase and trigger (reflexively) changes in  the speaker’s  output from normal to lombard speech, or cause the speaker to add in or switch to gestures as visual output.  Another change in external constraints may occur in the social context, such as when the number of people could increase from one to several, requiring the system to switch from spoken to written interaction in order to preserve privacy.

\subsection{Recommendations}
The range of modalities used in today’s dialog systems should be broadened, for example by including prosody, haptics, or motion capture as input modalities, or virtual or physical avatars or varying visual stimuli as output modalities. These additional modalities enrich our understanding of the users, improve user modeling by having a more holistic view of their behaviors, and augment their capability by creating new ways for users to interact with the machine.

By creating a broader range of modalities, there is a need to produce data sets covering these modalities and sharing them with the community to foster interdisciplinary and cross-institutional collaboration which is paramount as the complexity of the involved technologies grows.  Through the organization of open shared tasks as well as challenges and competitions held at conferences and workshops, multimodal dialog research will be further promoted, advanced, and accelerated.

Situated dialog should play a central role in future dialog systems research, especially considering the contributions of multimodality. For example, an exchange of physical objects between humans and an avatar as a coupled transaction requires not only spoken interaction, but also computer vision, recognition of objects and activities, fine-motor control, and situated reasoning.

The advances of incremental speech processing, for example incremental speech recognition, spoken language understanding, or speech synthesis, should be generalized to incremental multimodality. This is especially effective since several additional modalities, such as visual input, are continuous in nature, compared to spoken dialog which is often regarded as discrete, i.e. speaker-turn-based.

\section{Robust and Flexible Dialog Management}

\subsection{Current state of the art}

 In general, current intelligent agents target specific tasks and relevant user groups. They are quite brittle and inflexible, refusing to understand the user who is not aware of the commands the agent expects or of the type of language that the agent supports. 

Both industry and the research community have an interest in moving intelligent agents to multi-domain, multi-task, multi-topic conversational agents. But, perhaps surprisingly, even relatively simple tasks that require linking capabilities from two existing agents are hard to support. For example, consider a task as simple as “set a timer for cooking an egg”. 
\begin{itemize}
    \item User: Alexa, set a timer.
    \item Alexa: For how long?
    \item User: How long does it take to hard-boil an egg?
$\leftarrow$ To get this information, Alexa has to talk to Google search, find the amount of time, return the string from that search, reformulate it to a time, and fill the slot in the timer task) 
\end{itemize}

The challenge here is to interleave any structured task for which an explicit task structure exists with the results from search. There are many similar tasks, where constraints or information from multiple existing systems need to be integrated, such as wanting to check what the weather is going to be while booking a restaurant and deciding whether it will be possible to sit outside for dinner. 
Thus the steps in the process are:
\begin{itemize}
    \item combine an agent based on discrete information, with one based on search, which could retrieve any kind of answer.
    \item try to get two components to talk together
    \item integrate different data sources for the same task
\end{itemize}

Users often have slightly different goals or preferences than the ones that the agent is designed to support because they may not know what the underlying assumption is. This results in many limitations for the agent. For example, most natural language understanding components in dialog architectures assume that each utterance contains a single intent, so they may ignore additional information if the user mentions multiple intents in a single utterance.
To summarize:
\begin{itemize}
    \item current systems are very rigid
    \item it is hard to truly achieve mixed initiative
    \item at present the impression of mixed initiative is given by doing slot filling in different orders
\end{itemize}

\subsection{Desired flexibility}
Ideal systems should have more of the flexibility that people bring to a task-oriented dialogue. When people (e.g. a domain expert and a novice with a problem to solve) work together to find a solution, there is a natural sharing of initiative. Each presents contributions (e.g. candidate solutions or nuances or context of the problem) and the other provides feedback signals of attention, perception, understanding, and attitudinal reaction~\cite{allwood1992feedback}. Either participant can send signals of desire to speak, and can take initiative when they have something relevant to add. Many kinds of context are used including
\begin{itemize}
    \item What has been said recently
    \item The (mutually) perceivable or imaginable environment
    \item Mutual knowledge about the situation, including beliefs, desires, and plans
    \item Common-sense knowledge about likely eventualities and motivations.
\end{itemize}
These aspects of context allow for many convenient facets of dialogue that are largely missing from human-computer dialogue:
\begin{itemize}
    \item Allowing abbreviated or elliptical expressions to talk about salient entities and actions
    \item Allowing plan recognition to address the goals rather than specific mentioned actions, and provide helpful responses that even ignore a direct request, where it can be inferred that the requested action or information is not relevant to the goals.
\end{itemize}

Additionally, people often provide incremental and different sized contributions, where the full contribution is often shaped by feedback from the addressee while the contribution is being expressed. E.g. signals of lack of understanding may lead to elaboration, while acting on the contribution early may signal that a description is already good enough. Signals of interest may encourage elaboration of a topic, while signals of lack of strong interest may lead to either a change in topic or motivation for why the topic is relevant to the addressee.

Current work is just beginning to explore dialogs that cross domains. There is great potential to leverage assets from multiple kinds of agent activities or “skills”, knowledge resources, and social dialog to create a fuller experience, however more work is needed to realize this potential. For complex tasks, subdialogs should engage in simpler domains, but context and high-level goals must be maintained across these sub-dialogs for a coherent whole. Novel tasks should be achievable as a combination of simpler tasks.
Social dialog should be supported in that it contributes to building a relationship and common ground between system and user. This can appear both between and within tasks, often combining with other tasks within the same utterance.

\subsection{Recommendations}
\begin{itemize}
    \item Research should be highly interdisciplinary, including linguistics, pragmatics and reasoning
    \item Inference should be based on the user's goals and beliefs
    \item Structured representations that support inference should be combined with neural models dealing with unstructured text
    \item Research should be inspired by human dialog techniques in order to efficiently address others’ goals. Context should be preferred to making explicit representations
    \item Explicit representations of context should be designed in a way that enables them to  be shared with other domains, tasks and applications
    \item Agents should support different user style of interaction preferences
\end{itemize}

\section{Intelligent Agents as Good Actors}
\label{sec:ethics}
\subsection{Introduction}
Just as for artificial intelligence (AI) more generally, a host of ethical issues arise in research on intelligent agents. Research in this area will benefit from general advances in AI, but there are also special considerations for intelligent agents and potential new research directions that are highlighted in this section. 

\subsection{Ethics issues in intelligent agent research}

Since much of the current research on intelligent agents involves data-driven machine learning, the resulting systems are susceptible to: issues of bias in the collection of data (which leads to unintended bias in decisions); privacy issues associated with user data; and interpretability of system decisions. The issue of privacy is a concern for deep neural networks because of their ability to effectively memorize the data~\cite{carlini2019secret}. It is particularly challenging for spoken dialog systems because a user’s voice is their identity. Since users can now be identified from their voices, it is possible for their voices to be used for identity theft. This poses challenges for using cloud-based ASR services, as well as the storing and sharing of data. Cloud-based ASR gives better performance than on-device modules but requires transmitting the speech. Agent designers using cloud-based services should be aware of any potential commercial use of this data and inform users. A user-centric intelligent agent may also store information about its users to improve personalization (meta-data, usage patterns, and so on), which raises additional issues of data security. Privacy is of great concern for sharing both speech and text data associated with dialog, since there are currently no strong methods for de-identification.

Particular to intelligent agents is their interaction with human users, and so it is important for researchers to carefully consider how the system can influence individual behavior and mental state. In many applications, it is important for users to develop a level of trust in the system, and research on interaction strategies to that end is important. However, since the system may make errors, it is important that it be able to communicate uncertainty in an effective way. Passing the Turing Test may not be an appropriate objective – the user should know when they are talking to a bot. Technology designed to deceive humans is much more susceptible to use in phishing scenarios or other types of user manipulation. Another issue that arises during interaction with users is the possibility of unintentionally providing incorrect information or poor advice; potential problems need to be considered and the cost of such mistakes accounted for in system design. 

Government support of intelligent agent research should focus on societal needs and challenges. Like any technology, intelligent agents can be used for good or bad purposes. System designers should consider the potential for a system to be used for destructive purposes, such as hate speech, bullying or stalking. When talking to
%JH: not sure what the next bit means
a bot, some users are uninhibited and interactions can be highly toxic or offensive, which systems must learn to handle appropriately. Lastly, dialog research needs to develop mechanisms for robustness to adversarial attack, which may be in the form of biasing the data used for learning or injecting false information into a knowledge base or backend database.

Some ethical issues are dependent on the user population and/or application scenario, but all dialog system research will involve some subset of these issues. As such, it is important for educational activities associated with research to include training in ethical thinking. Furthermore, projects with outreach activities should incorporate discussion of ethical issues into their programs to raise public awareness.

\subsection{Balancing Tensions Between User Concerns and Research Advances}
It is important to recognize that fully addressing these concerns limits opportunities for research. For example, the scientific objective of research reproducibility can be inhibited by failure to share data; however, it is very difficult to share data and also to preserve privacy, as noted above. Using paid crowd-workers rather than users with real needs may avoid privacy concerns, but questions are then often raised about the validity of the data, given that workers may accept undesirable results to complete their tasks more quickly. Using simulated users avoids exposing real users to system errors in reinforcement learning, but it limits the potential to understand user variability. When privacy concerns exclude the possibility of looking at data or listening to speech signals, it can be much harder to diagnose and improve systems, resulting in preserving or even reinforcing biases in system performance. If only text transcripts of speech are available and audio access is restricted, then systems cannot explore the use of prosodic cues to engagement, sentiment, intent, and so on. There is also a conflict between user modeling and the potential for bias and restriction of access to information; personalization can be a double-edged sword. Discarding data for privacy reasons eliminates the potential for longitudinal studies. The problem of toxic language on the web (and in user interactions) creates a need for detecting toxicity to avoid introducing it into system behavior, but this creates the problem of exposing unpleasant material to researchers who are annotating or analyzing data.

These tensions underscore the importance of ethical training for dialog researchers and the need to inform users about both risks and benefits. They also create a need for creative thinking and innovation in data collection and system evaluation to support some level of reproducibility in dialog research.

\subsection{Recommendations}
Fundamental research is needed to preserve the integrity of future dialog agents and the security of human users. It is important to recognize that future research should study not only designing better systems, but also creating methods to protect the data. Therefore, we divide the research directions into two folds: ethic-related research for well-behaved agents, and privacy preserving methods on data.

For the data side, an important research venue is developing privacy-preserving algorithms that can remove sensitive information from raw data, including speech, visual, language process and machine learning. This is a recently emerging research direction due to its significant social implications since machine learning systems are sometimes adopted into commercial systems. A challenge for this research direction include the tension between the degree of anonymization vs the degree of information loss. 

On the other hand, for the system/agent side, the first important research direction is to develop robust detection methods to automatically recognize inappropriate behavior, including offensive language, misleading information, discrimination against certain populations etc. This requires a joint effort in the creation of related datasets and shared tasks and the development of machine learning methods that can solve this challenge. A second direction on the system side is to develop dialog agents that are resilient to adversarial users. Take the Microsoft Tay as an example. Its self-learning mechanism from user feedback from real users turned out to be a disaster, i.e. there is a significant amount of adversarial input from human users that made Microsoft's Tay post inflammatory and offensive tweets through its Twitter account, causing Microsoft to shut down the dialog service only 16 hours after its launch. That said, continuous learning from real interaction data is crucial for agent behavior, e.g. personalization, reinforcement learning etc, the best solution is to take the potential adversarial into account when designing future algorithms. 

Along with the progress in detecting inappropriate behavior and intelligent agent evaluation that this report recommends for the next 5-10 years, there will be a need for standardization, i.e. the development of a ''drivers license'' to comprehensively evaluate the quality of an intelligent agent before it can be deployed in production.

\subsection{Research-adjacent recommendations}
\begin{itemize}
    \item Research plans within scientific proposals should raise and address at least one ethical issue.
    \item Education of students on ethical issues should be an important topic.
    \item Develop outreach efforts for increasing public awareness in collaboration with target user demographics.
    \item For every NSF/government proposal or grant that includes the release of data, the data should have a data sheet~\cite{gebru2018datasheets} or similar.
\end{itemize}

\section{Conclusions}

Intelligent agents of the future will adapt to the wishes of the user. They will be able to converse with the user in a large variety of situations, through a wide variety of modalities and in a secure environment.

The USER Workshop participants recommend that the intelligent agent community attend to the user in all of its research; that its agents adapt to the individual user’s style, verbosity, goals and beliefs.

Research teams should be multidisciplinary, including specialists from domains concerning the agent itself as well as its potential application.

Intelligent agents must deal with ever-increasingly complex situations including more modalities of interaction, multiparty dialogs, action on physical objects, longer dialogs and more frequent use by the same users.

The research infrastructure should include open source robust and reusable tools, well-documented pre-training models and novel machine learning methods that can deal with non-annotated data. All data collected should be shared freely with the whole community.

The research community must create shared tasks that define scientific goals and measure resulting progress. The assessment of intelligent agents should take place within an interactive environment. Common research platforms that satisfy these needs should be encouraged.

The intelligent agent and its creators must be good actors, protecting user privacy through the development of dedicated algorithms and through the automatic detection of inappropriate behavior. For this, research plans within scientific proposals should raise and address at least one ethical issue. Researchers should be encouraged to educate students on ethical issues and to develop public awareness outreach efforts. 

%In conclusion, the USER Workshop participants suggest that the intelligent agent community pursue research in seven broad areas: applications, infrastructure, dynamic user-agent interaction, dealing with low resource conditions, multimodal, grounded and situated interaction, robust and flexible dialog management and intelligent agents as good actors. 
%There were six crosscutting themes found in many of the areas. These recurring themes reveal the strong agreement of the participants on intelligent agents' future directions. These themes should underlie future research and should be strongly encouraged by the National Science Foundation. All areas agreed that research should be user-centric. Research should concern system architecture, especially plug-and-play systems for ease of access for newcomers. Research teams should be multidisciplinary. Researchers should attend to creating agents that can deal with multiple conditions and situations. In order to spark progress and to measure it, there should be shared tasks. And every research project should take ethics and privacy into account. 

\bibliography{bib.bib}
\bibliographystyle{acl_natbib}

\end{document}